\documentclass[journal]{IEEEtran}

\usepackage[T1]{fontenc}
\usepackage{times}
\usepackage{epsfig}
\usepackage{graphicx}
\usepackage{amsmath}
\usepackage{amssymb}
\usepackage{subcaption}
\usepackage{epstopdf}
\usepackage{tikz}
\usepackage{hyperref}
\usetikzlibrary{fit,positioning}
\usetikzlibrary{shapes.geometric,positioning}
\usepackage{algorithm}
\usepackage{algorithmic}

\ifCLASSINFOpdf
\else
\fi

\begin{document}

\title{Accelerometer based Activity Classification with Variational Inference on Sticky HDP-SLDS}

\author{Mehmet~Emin~Basbug,~\IEEEmembership{Student~Member,~IEEE,}
        Koray~Ozcan,~\IEEEmembership{Student~Member,~IEEE,}
        Burak~Kakillioglu,~\IEEEmembership{Student~Member,~IEEE,}
        and~Senem~Velipasalar,~\IEEEmembership{Senior~Member,~IEEE}
\thanks{M. E. Basbug is with the Department of Electrical  Engineering, Princeton University, Princeton, NJ, 30332 USA e-mail: mbasbug@princeton.edu}
\thanks{K. Ozcan, B. Kakillioglu, and S. Velipasalar are with the Dept. of Electrical Engineering and Computer Science, Syracuse University, Syracuse,
NY, 13244 USA. e-mail: \{kozcan, bkakilli, svelipas\}@syr.edu.}
\thanks{This work has been funded in part by National Science Foundation (NSF) CAREER grant CNS-1206291 and NSF Grant CNS-1302559. }
}

\markboth{IEEE Person-Centered Signal Processing for Assistive, Rehabilitative and Wearable Health Technologies, September~2015}
{Shell \MakeLowercase{\textit{et al.}}: Bare Demo of IEEEtran.cls for Journals}

\maketitle

\begin{abstract}
As part of daily monitoring of human activities, wearable sensors and devices are becoming increasingly popular sources of data. With the advent of smartphones equipped with acceloremeter, gyroscope and camera; it is now possible to develop activity classification platforms everyone can use conveniently. In this paper, we propose a fast inference method for an unsupervised non-parametric time series model namely variational inference for sticky HDP-SLDS(Hierarchical Dirichlet Process Switching Linear Dynamical System). We show that the proposed algorithm can differentiate various indoor activities such as sitting, walking, turning, going up/down the stairs and taking the elevator using only the acceloremeter of an Android smartphone Samsung Galaxy S4. We used the front camera of the smartphone to annotate activity types precisely. We compared the proposed method with Hidden Markov Models with Gaussian emission probabilities on a dataset of 10 subjects. We showed that the efficacy of the stickiness property. We further compared the variational inference to the Gibbs sampler on the same model and show that variational inference is faster in one order of magnitude.
\end{abstract}

\begin{IEEEkeywords}
Smartphone, accelerometer, wearable sensor, activity classification, unsupervised learning, variational inference, nonlinear timeseries
\end{IEEEkeywords}

%
\IEEEpeerreviewmaketitle

\section{Introduction}

\IEEEPARstart{A}{ccurate} and effective activity monitoring is becoming more important to reach higher quality of living. Current smart phones and tablets, equipped with powerful processors and a wide variety of sensors, have become ideal platforms for activity monitoring. Especially for elderly people, accurate monitoring of daily activities is ultimately important for ovearall health condition analysis. ~\cite{Zhang} describe a hierarchical method of activity classification based on a smart phone, equipped with an embedded 3D-accelerometer, worn on a belt. With two multi-class SVM classifiers for rule based reasoning of motion and motionless activities, they were able to achieve an accuracy of $82.8\%$ for six different activities. Accelerometers have also been used in the activity classification for sports activities~\cite{Taylor} in order to capture and archive various training statistics. One of the key points discussed in~\cite{Taylor} is that the placement of the phone makes it hard to generalize a model trained offline. This is the main drawback of the supervised approach. Ideally, an adaptive model should be able to learn the activities on the fly and expands its repertoire when it encounters a new type of activity. In other words, we need an unsupervised non-parametric model.

There are a few unsupervised learning methods applied to wearable accelerometer data. For instance, \cite{ladha2013dog} applied k-nearest neighbor algorithm to the problem of activity classification for dogs with $70\%$ classification accuracy.\cite{kwon2014unsupervised} used unsupervised learning applied on sensor data including accelerometer, magnetometer and gyroscope for smartphones to classify human activities. They used a method based on hierarchical clustering coupled with Gaussian Mixture Models. \cite{trabelsi2013unsupervised} proposed unsupervised human activity classification using Hidden Markov Model(HMM) in a multiple regression context using sensors attached to ankle, hip and chest. The sensors are used to record three-dimensional accelerometer data resulting in nine-dimensional data for classifier methods. Moreover, \cite{taniguchi2011unsupervised} used a sticky hierarchical Dirichlet process HMM (sticky HDP-HMM) to segment human activities such as waving good-bye, walking and throwing a ball so that a robot can imitate them. They measure angles of various joints of human body to estimate unknown number of human activities in order to classify them. In all these methods, the autocorrelated nature of the accelerometer data is not accounted for. Most notably HMM based approaches are making the assumption that each observation is i.i.d. given the activity. This is valid approach if the activities differ in the mean of the observation but it fails if the activities differ mostly in the covariance of the observation.

To collect the accelerometer data, the smartphone is attached to a belt around the waist. There are two benefits of this placement i) it is close to the center of gravity therefore the images from the camera are more stable~\cite{Kangas2008} ii) it has been recently shown that the acceloremeter placed around the waist provides the highest accuracy of classification~\cite{atallah2011sensor}. In terms of activities, the subjects are instructed to choose among walking, going up and down the stairs, turning, taking the elevator, sitting and lying down in a random order. The subjects were also encouraged to repeat each activity random amount of times.

Activity classification problem exhibits a piecewise linear autocorrelation characteristic. One of the popular models for such timeseries data is the switching linear dynamical system (SLDS) model. SLDS model consists of a collection of linear dynamical system (LDS) models which are primarily used to infer the hidden dynamical behavior of a noisy system from noisy observations \cite{kalman1960new}. SLDS models have successfully been used to model the regime switching in interest rates \cite{gray1996modeling}, the human motion from video data \cite{pavlovic2000learning}, the dance of honey bees \cite{fox2008nonparametric}, the respiration pattern of a sleep apnea patient \cite{ghahramani2000variational}, the interconnectivity of brain regions \cite{smith2010identification}, and many others.

SLDS can be described as a hybrid of hidden Markov model (HMM) and LDS model. The linear Gaussian process models the dynamical behavior of the system within each temporal mode and the hidden Markov chain captures the sequence of temporal modes. Each hidden state in the Markov chain corresponds to a distinct temporal mode and has its own parameters for the linear Gaussian process it governs. In an ordinary SLDS, the number of temporal modes must be specified ahead of time; however, this information may not always be available. Furthermore, it might be desirable to expand the repertory of the model as new data arrives. With Bayesian nonparametric techniques, this problem can be alleviated. Letting a Hierarchical Dirichlet Process (HDP) determine the state transition, one can allow for countably infinite number of states. This method is first explored for HMMs in \cite{beal2002infinite}. The initial attempts resulted in oscillatory behavior in practice. \cite{fox2008hdp} modified the generative process to ensure the mode persistence. Later this concept is extended to SLDS in \cite{fox2011sticky} where the new model is named sticky HDP-SLDS.

The Bayesian approach to HMM, LDS and SLDS models has been explored extensively. A variational inference algorithm is developed for a fully hierarchical LDS model with automatic relevance determination (ARD) in \cite{beal2003variational}. \cite{ghahramani2000variational}, \cite{pavlovic2000impact} and \cite{oh2005variational} proposed variational inference algorithms for various SLDS models. These early works do not have the non-parametric treatment of the state transition. \cite{fox2011sticky} incorporated HDP into SLDS; however, they used Gibbs sampling for inference with the assumption that each temporal mode shares the same observation noise model. The use of Gibbs sampling degraded the speed of inference significantly.

The major contribution of this work is to develop a variational inference algorithm for the sticky HDP-SLDS model presented in \cite{fox2008nonparametric}. We restrict ourselves to ARD modeling; however, we allow for each temporal mode to have a distinct sensor model. We further extend the original model to have mutliple observations sharing the same underlying switching beahvior. We compare the original Gibbs sampling and variational inference on synthetic data in their ability to capture the true temporal mode sequence as well as their speed.

Having shown the superiority of variational inference for the inference speed, we employed it for the activity classification problem. Previously, \cite{zhu2011bayesian} used sticky HDP-HMM on a similar dataset with a limited number of activities. We experimentally verified that sticky HDP-SLDS with variational inference is a fast and accurate unsupervised method suitable for the activity classification with a single accelerometer sensor.

\section{Background}
\subsection{Switching Linear Dynamical Systems}
\label{skfm}

An LDS model with an underlying state $x_{t} \in \mathbb{R}^{d_{x}}$ and observation $z_{t} \in \mathbb{R}^{d_{z}}$ can be described as
\begin{align*}
x_{t+1} &= F x_{t} + v_{t}&&v_{t}\sim N(0,U)\\
z_{t} &= H x_{t} + w_{t}&&w_{t}\sim N(0,R)
\end{align*}
where $F_{d_{x} \times d_{x}}$ is the state dynamics matrix, $H_{d_{z} \times d_{x}}$ is the observation matrix, $U_{d_{x} \times d_{x}}$ and $R_{d_{z} \times d_{z}}$ are the state and observation noise covariance matrices respectively. \cite{beal2003variational} showed that without loss of generality we can assume $U_{d_{x} \times d_{x}}$ to be the identity matrix $I_{d_{x} \times d_{x}}$.

SLDS requires another state variable $s_{t} \in \mathbb{Z}^{+}$ for the Markov chain of temporal modes. An SLDS with $N$ observations sharing the same temporal mode sequence can be described as
\begin{align*}
s_{t+1} \mid s_{t} &\sim \pi_{s_{t}}\\
x_{t+1}^{n} &= F_{s_{t}} x_{t}^{n} + v_{t},&&v_{t}\sim N(0,I)\\
z_{t}^{n} &= H_{s_{t}} x_{t}^{n} + w_{t},&&w_{t}\sim N(0,R_{s_{t}})
\end{align*}
where $\pi$ is the transition matrix and the initial states are given by $s_{1}\sim \pi_{0}$ and $x_{1}^{n}\sim N(\mu_{s_{t}},\Sigma_{s_{t}})$.

\subsection{Hierarchical Dirichlet Process}
\label{hdp}

A two level HDP can be described as a collection of Dirichlet Processes (DP) $\left \{ \mathcal{G}_{j}\right \}_{j=1}^{\infty}$ each characterized by the parameter $\alpha$ and the base measure $\mathcal{G}_{0}$ which is also drawn from a DP with the parameter $\gamma$ and the base measure $\mathcal{H}$. Mathematically we write
\begin{align*}
\mathcal{G}_{0} &\sim DP(\gamma,\mathcal{H}),&&\mathcal{G}_{j} \sim DP(\alpha,\mathcal{G}_{0})
\end{align*}
To construct such HDP, \cite{wang2011online} used Sethuraman's stick breaking procedure as follows
\begin{align}
\bar{\beta}_{i} &\sim Beta(1,\gamma)&&\beta_{i} = \bar{\beta}_{i}\prod_{i'=1}^{i-1}(1-\bar{\beta}_{i'})\label{gem_beta}\\
\Theta_{i} &\sim \mathcal{H}&& \mathcal{G}_{0} = \sum_{i=1}^{\infty}\beta_{i}\delta_{\Theta_{i}}\notag\\
\bar{\pi}_{ii'}&\sim Beta(1,\alpha)&&\pi_{ii'}' = \bar{\pi}_{ii'}\prod_{i''=1}^{i'-1}(1-\bar{\pi}_{ii''})\label{gem_pi'}\\
\Phi_{ii'}&\sim \mathcal{G}_{0}&&\mathcal{G}_{i} = \sum_{i'=1}^{\infty}\pi_{ii'}'\delta_{\Phi_{ii'}}\;\;\;.\notag
\end{align}
In the case of SLDS, $\Theta_{i}$ refers to the set of parameters of $i^{th}$ temporal mode. $\mathcal{G}_{0}$ acts as an average distribution over the temporal modes. $\mathcal{G}_{j}$ is related to the transition probabilities from $j^{th}$ temporal mode. The connection between $\Phi_{ii'}$ and $\Theta_{j}$ can be established with a set of indicator variables $\{c_{ii'}\}_{i'=1}^{\infty}$ such that $\Phi_{ii'} = \Theta_{c_{ii'}}$ and $c_{ii'} \sim Mult(\beta)$. Now we can write the transition probabilities as
\begin{align}
\pi_{ij} = \sum_{i' = 1}^{\infty}\pi_{ii'}'\mathbf{1}\{c_{ii'}=j\}
\label{mapping_pi}
\end{align}
With this construction we ensure that the transitioned state shares the same base measure $\mathcal{G}_{0}$ in expectation i.e. $\mathbb{E}[ \pi_{ij} \mid \beta ]= \beta_{j}.$ As noted earlier, this implementation might fail to capture the mode persistency. To solve this problem, \cite{fox2011sticky} introduced the concept of sticky HDP. For the sticky version we make the following modifications
\begin{align*}
\bar{\pi}_{ii'}&\sim Beta(1,\alpha + \kappa)& \text{and} && c_{ii'} \sim Mult(\frac{\alpha\beta + \kappa \delta_{i}}{\alpha+\kappa})
\end{align*}
where $\kappa$ is the self transition parameter. With this construction we obtain a slightly different expectation for the transitioned state
\begin{align*}
\mathbb{E}[ \pi_{ij} \mid \beta, \kappa, \alpha] = \frac{\alpha}{\alpha+\kappa}\beta_{j}+\frac{\kappa}{\alpha+\kappa}\mathbf{1}\{i=j\}.
\end{align*}

\subsection{Variational Inference}
\label{variational}
In a well constructed graphical model, the latent variables encode the hidden pattern in the data and it is feasible to learn this pattern by computing the posterior of latent variables given observations \cite{blei2014build}. Using chain rule we write
\begin{align*}
P(\mathbf{L} \mid \mathbf{Z}) = \frac{P(\mathbf{Z},\mathbf{L})}{P(\mathbf{Z})}
\end{align*}
where $\mathbf{L} = \{L_{1},L_{2},\ldots\}$ and $\mathbf{Z} = \{Z_{1},Z_{2},\ldots\}$ are the sets of latent variables and observed variables respectively. The inference of the latent variables is relatively easy given $P(\mathbf{Z}) = \int_{\mathbf{L}}P(\mathbf{Z},\mathbf{L})d\mathbf{L}$. However, this integration becomes intractable for many practical models. With no analytical solution, one has to resort to approximate inference methods. A common approach is to use sampling techniques such as Markov Chain Monte Carlo (MCMC) for the inference \cite{fox2008nonparametric}.  An alternative strategy is to use divergence methods such as variational inference. In variational inference, the probability of observation is approximated by another set of distributions over the latent variables. Using Jensen's inequality we write
\begin{align}
\ln P(\mathbf{Z}) =& \ln\int_{\mathbf{L}}P(\mathbf{Z},\mathbf{L})\frac{Q(\mathbf{L})}{Q(\mathbf{L})}d\mathbf{L}\notag\\
\geq& \mathbb{E}_{Q(\mathbf{L})}[\ln P(\mathbf{Z},\mathbf{L})] - \mathbb{E}_{Q(\mathbf{L})}[\ln Q(\mathbf{L})]\notag\\
=&ELBO\label{elbo_initial}
\end{align}
where $Q(\mathbf{L})$ is the variational distribution used for approximation and $ELBO$ is the evidence lower bound. \cite{jordan1999introduction} showed that maximizing $ELBO$ is equivalent to minimizing the KL divergence between the posterior distribution $P(\mathbf{L}\mid\mathbf{Z})$ and the variational distribution $Q(\mathbf{L})$.

To simplify further, we make the mean field assumption which requires all the variational distributions to be in the mean field variational family. In this family, each latent variable is independent and governed by its own parameter, $Q(\mathbf{L}) = \prod_{i} Q(L_{i} \mid \lambda_{i}).$ With this assumption, we posit the inference as an optimization problem which can be solved using a coordinate ascent algorithm. Furhermore, it gives the flexibility of using stochastic methods and scalability \cite{hoffman2013stochastic}. The update for each variational distribution can be represented as
\begin{align}
Q(L_{i}) = \frac{1}{c_{i}}\exp \left\{\mathbb{E}_{Q(\sim L_{i})}[\ln P(\mathbf{Z},\mathbf{L})]\right\}\label{general_var_update}
\end{align}
where $c_{i}$ is the normalization constant and the expectation is taken under $Q(\sim L_{i}) \doteq \prod_{j\neq i} Q(L_{j})$.

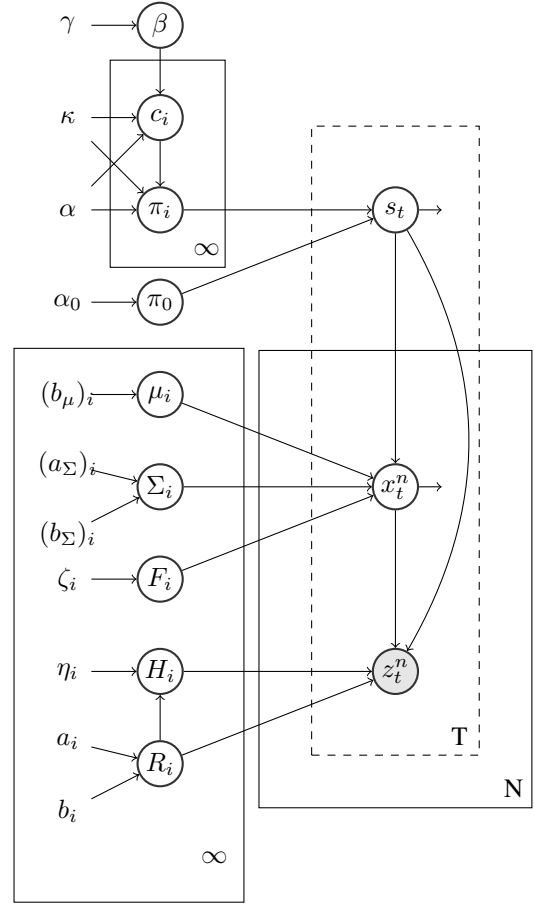
\begin{figure}[t]
  \centering
  \begin{tikzpicture}
    \tikzstyle{main}=[circle, minimum size = 6mm, thick, draw =black!80, node distance = 6mm]
    \tikzstyle{connect}=[-latex, thick]
    \tikzstyle{main2}=[circle, minimum size = 6mm, thick, draw =black!80, node distance = 25mm]
    \tikzstyle{connect}=[-latex, thick]
    \tikzstyle{main3}=[circle, minimum size = 6mm, thick, node distance = 3mm]
    \tikzstyle{connect}=[-latex, thick]
    \tikzstyle{hyper}=[rectangle, minimum size = 6mm, thick, node distance = 6mm]
    \tikzstyle{connect}=[-latex, thick]
    \tikzstyle{hyper2}=[rectangle, minimum size = 6mm, thick, node distance = 3mm]
    \tikzstyle{connect}=[-latex, thick]
    \tikzstyle{box}=[rectangle, draw=black!100]
      \node[main, fill = white!100] (beta) [label=center:$\beta$] { };
      \node[main] (ci) [below=of beta,label=center:$c_{i}$] { };
      \node[main] (pi) [below=of ci,label=center:$\pi_{i}$] { };
      \node[main] (pi_zero) [below=of pi,label=center:$\pi_{0}$] { };
      \node[main] (mu) [below=of pi_zero, label=center:$\mu_{i}$] { };
      \node[main] (Sigma) [below=of mu, label=center:$\Sigma_{i}$] { };
      \node[main] (F) [below=of Sigma, label=center:$F_{i}$] { };
      \node[main] (H) [below=of F, label=center:$H_{i}$] { };
      \node[main] (R) [below=of H, label=center:$R_{i}$] { };
      \node[hyper] (gamma) [left=of beta,label=center:$\gamma$] { };
      \node[hyper] (kappa) [left=of ci,label=center:$\kappa$] { };
      \node[hyper] (alpha) [left=of pi,label=center:$\alpha$] { };
      \node[hyper] (alpha_zero) [left=of pi_zero,label=center:$\alpha_{0}$] { };
      \node[hyper] (b_mu) [left=of mu,label=center:$(b_{\mu})_{i}$] { };
      \node[hyper2] (a_sigma) [below=of b_mu,label=center:$(a_{\Sigma})_{i}$] { };
      \node[hyper2] (b_sigma) [below=of a_sigma,label=center:$(b_{\Sigma})_{i}$] {};
      \node[hyper] (zeta) [left=of F,label=center:$\zeta_{i}$] { };
      \node[hyper] (eta) [left=of H,label=center:$\eta_{i}$] { };
      \node[hyper2] (a) [below=of eta,label=center:$a_{i}$] { };
      \node[hyper2] (b) [below=of a,label=center:$b_{i}$] { };
      \draw [->] (gamma) -- (beta) ;
      \draw [->] (kappa) -- (ci) ;
      \draw [->] (kappa) -- (pi) ;
      \draw [->] (alpha) -- (ci) ;
      \draw [->] (beta) -- (ci) ;
      \draw [->] (ci) -- (pi) ;
      \draw [->] (alpha) -- (pi) ;
      \draw [->] (alpha_zero) -- (pi_zero) ;
      \draw [->] (b_mu) -- (mu) ;
      \draw [->] (a_sigma) -- (Sigma) ;
      \draw [->] (b_sigma) -- (Sigma) ;
      \draw [->] (zeta) -- (F) ;
      \draw [->] (eta) -- (H) ;
      \draw [->] (a) -- (R) ;
      \draw [->] (b) -- (R) ;
      \draw [->] (R) -- (H) ;
      \node[main2] (s) [right=of pi,label=center:$s_{t}$] { };
      \node[main2] (x) [right=of Sigma,label=center:$x_{t}^{n}$] { };
      \node[main3] (s_next) [right=of s] { };
      \node[main3] (x_next) [right=of x] { };
      \node[main2,fill = black!10 ] (z) [right=of H,label=center:$z_{t}^{n}$] { };
      \draw [->] (pi) -- (s) ;
      \draw [->] (pi_zero) -- (s) ;
      \draw [->] (mu) -- (x) ;
      \draw [->] (Sigma) -- (x) ;
      \draw [->] (F) -- (x) ;
      \draw [->] (H) -- (z) ;
      \draw [->] (R) -- (z) ;
      \draw [->] (s) -- (x) ;
      \draw [->] (x) -- (z) ;
      \draw [->] (s) -- (s_next) ;
      \draw [->] (x) -- (x_next) ;
      \draw [bend left,->] (s) to (z) ;
      \node[rectangle, inner sep=0mm, fit= (ci) (pi),label=south east:$\infty$] {};
      \node[rectangle, inner sep=4.5mm,draw=black!100, fit= (ci) (pi),xshift=1mm] {};
      \node[rectangle, inner sep=1mm, fit= (b_mu) (b) (mu) (R),label=south east:$\infty$] {};
      \node[rectangle, inner sep=6mm,draw=black!100, fit= (b_mu) (b) (mu) (R),yshift=-3mm,xshift=2mm] {};
      \node[rectangle, inner sep=3mm,fit= (s) (x) (z),label=south east:T] {};
      \node[rectangle, inner sep=8mm,draw=black!100,dashed, fit= (s) (x) (z)] {};
      \node[rectangle, inner sep=10mm,fit= (x) (z),label=south east:N] {};
      \node[rectangle, inner sep=15mm,draw=black!100, fit= (x) (z)] {};
  \end{tikzpicture}
  \caption{Graphical model for sticky HDP-SLDS. It should be noted that $\pi_{0}$ only affects $s_{1}$. Similarly $\mu_{i}$ and $\Sigma_{i}$ are only connected to $x_{1}^{n}$.}
  \label{hdpslds_graph}
\end{figure}

\section{Variational Inference for HDP-SLDS}
\label{var_hdpslds}
The graphical model for the multi observation sticky HDP-SLDS is depicted in Figure~\ref{hdpslds_graph}. $\{x_{t}^{n}\}_{t=1}^{T}$ is the hidden Markov process for $n^{th}$ observation sequence $\{z_{t}^{n}\}_{t=1}^{T}$ and  $\{s_{t}\}_{t=1}^{T}$ is the hidden Markov chain for temporal modes as described in Section~\ref{skfm}. $\pi_{i}$ and $c_{i}$ govern the transition from $i^{th}$ temporal mode characterized by $\Theta_{i} = \left \{\mu_{i},\Sigma_{i},F_{i},H_{i},R_{i} \right \}$. $\beta$ is related to the average distribution in Section~\ref{hdp}. $\pi_{0}$ is the distribution of the initial state $s_{1}$. The joint likelihood of the model can be written as
\begin{align}
&\ln P(\bar{\pi}_{0},\bar{\pi},\bar{\beta},C,\Theta,S,X,Z)\notag\\
&= \sum_{i = 1}^{\infty}\ln P(\bar{\pi}_{0i}\mid \alpha_{0}) +\sum_{i = 1}^{\infty}\sum_{i' = 1}^{\infty}\ln P(\bar{\pi}_{ii'}\mid \alpha,\kappa) \notag\\
&+\sum_{i = 1}^{\infty}\ln P(\bar{\beta}_{i}\mid \gamma) + \sum_{i = 1}^{\infty}\sum_{i' = 1}^{\infty}\ln P(c_{ii'}\mid \bar{\beta},\kappa,\alpha)  \notag\\
&+\sum_{k = 1}^{\infty}\ln P(\mu_{k}\mid b_{\mu}^{k}) + \ln P(\Sigma_{k}\mid a_{\Sigma}^{k},b_{\Sigma}^{k}) +\ln P(F_{k}\mid \zeta_{k})\notag\\
&+\sum_{k = 1}^{\infty}\ln P(R_{k}\mid a_{k},b_{k})+ \ln P(H_{k}\mid R_{k},\eta_{k})\notag\\
&+\ln P(s_{1} \mid \bar{\pi}_{0}) + \sum_{t=2}^{T} \ln P(s_{t} \mid s_{t-1},C,\bar{\pi})\notag
\end{align}
\begin{align}
&+ \sum_{n = 1}^{N}\left \{\ln P(x_{1}^{n}\mid s_{1},\mu,\Sigma)+ \sum_{t= 2}^{T} \ln P(x_{t}^{n}\mid x_{t-1}^{n},s_{t},F)\right.\notag\\
&\left. +\sum_{t = 1}^{T} \ln P(z_{t}^{n}\mid x_{t}^{n},R,H)\right \}
\label{joint_log}
\end{align}
With the mean field assumption, the variational distribution $Q(\bar{\pi}_{0},\bar{\pi},\bar{\beta},C,\Theta,S,X)$ can be decomposed into its components. Using the stick breaking construction described in Section~\ref{hdp} and ARD modeling, the variational distributions for the time-invariant latent variables can be written as
\begin{align*}
Q(\bar{\beta}) =& \prod_{i=1}^{K-1} Beta(\bar{\beta}_{i} \mid u_{i}^{\beta},v_{i}^{\beta})\\
Q(\bar{\pi}_{0}) =& \prod_{i=1}^{K-1} Beta(\bar{\pi}_{0i} \mid u_{0i}^{\pi},v_{0i}^{\pi})\\
Q(\bar{\pi}) =& \prod_{i=1}^{K}\prod_{i'=1}^{K-1} Beta(\bar{\pi}_{ii'} \mid u_{ii'}^{\pi},v_{ii'}^{\pi})\\
Q(C) =& \prod_{i=1}^{K}\prod_{i'=1}^{K} Mult(c_{ii'} \mid \phi_{ii'})\\
Q(\mu) =& \prod_{i=1}^{K} \prod_{d=1}^{d_{x}} N(\mu_{id} \mid \mu_{id}^{\mu},\Sigma_{id}^{\mu})\\
Q(\Sigma^{-1}) =& \prod_{i=1}^{K} \prod_{d=1}^{d_{x}}  Ga(\sigma_{id} \mid a_{id}^{\sigma},b_{id}^{\sigma})\\
Q(F) =& \prod_{i=1}^{K}\prod_{d=1}^{d_{x}} N(f_{id} \mid \mu_{id}^{F},\Sigma_{id}^{F})\\
Q(R^{-1}) =& \prod_{i=1}^{K}\prod_{d=1}^{d_{z}} Ga(\rho_{id} \mid a_{id}^{\rho},b_{id}^{\rho})\\
Q(H) =& \prod_{i=1}^{K}\prod_{d=1}^{d_{z}} N(h_{id} \mid \mu_{id}^{H},\rho_{id}^{-1}\Sigma_{i}^{H})
\end{align*}
where $\Sigma_{i}^{-1} = diag(\sigma_{i})$, $R_{i}^{-1} = diag(\rho_{i})$, $f_{id}$ refers to the $d^{th}$ row of $F_{i}$, $h_{id}$ refers to the $d^{th}$ colummn of $H_{i}$ and $K$ is the truncation parameter i.e.
\begin{align*}
&Q(\bar{\beta}_{K}=1) = 1 \\
&Q(\bar{\pi}_{0K}=1) = 1\\
&Q(\bar{\pi}_{iK}=1) = 1 \text{\;\;for\;\;} i=1,\ldots,K
\end{align*}
The truncation parameter $K$ might give the impression that we are working with $\Theta_{i}$ to grow as needed. $K$ is simply the upper bound on the number of possible modes \cite{wang2011online}.

We can make use of the conjugacy relations to update the time-invariant variational parameters \cite{hoffman2013stochastic}. The details of these updates are given in Table~\ref{var_params}. We will explain the update for $\phi_{ii'}$ in further detail as it requires an approximation. We note that $\mathbb{E}[\beta_{k}]\footnote{For the rest of the paper every expectation is taken under the variational distribution Q unless it is specified explicitly. The reader is expected to identify the relevant components of Q.},\mathbb{E}[\beta_{k}^{2}]$ and $\mathbb{E}[\ln\beta_{k}]$ are easy to compute since $\left \{\bar{\beta_{k}}\right \}_{k=1}^{K}$ are i.i.d. ~\cite{wang2011online}. The prior value of $\phi_{ii'}$ is given by
\begin{align*}
\phi_{ii'}(k) = \frac{\alpha\mathbb{E}[\beta_{k}] + \kappa \mathbf{1}\{i=k\}}{\alpha+\kappa}
\end{align*}
The posterior for $\phi_{ii'}(k)$ is proportional to
\[
 \exp
  \begin{cases}
    \ln\alpha+\mathbb{E}[\ln\beta_{k}] + U^{\phi} & \text{if } i \neq k \\
    \ln\kappa+\mathbb{E}[\ln(1+\frac{\alpha\beta_{k}}{\kappa})] + U^{\phi} & \text{if } i=k .
  \end{cases}
\]
where $U^{\phi} = \mathbb{E}[\ln \pi_{ii'}']\sum_{t=2}^{T}\mathbb{E}[s_{t-1}(i),s_{t}(k)]$

Using the approximations
\begin{align*}
\ln(1+y) &\approx y - 0.5y^2, &\qquad{} 0<y<1\\
\ln(1+y) &\approx \ln y +1/y, &\qquad{} 1<y\\
\mathbb{E}[1/y] &\approx \mathbb{E}[y^{2}]/\mathbb{E}[y]^{3}
\end{align*}
we write $\mathbb{E}[\ln(1+\frac{\alpha\beta_{k}}{\kappa})] $ as
\[\approx
  \begin{cases}
    \frac{\alpha}{\kappa}\mathbb{E}[\beta_{k}]-\frac{\alpha^{2}}{2\kappa^{2}}\mathbb{E}[\beta_{k}^{2}] & \text{if } \kappa > \alpha \mathbb{E}[\beta_{k}] \\
    \ln\alpha -\ln\kappa+\mathbb{E}[\ln\beta_{k}]+ \frac{\kappa}{\alpha}(\frac{\mathbb{E}[\beta_{k}^{2}]}{\mathbb{E}[\beta_{k}]^{3}}) & \text{if } \kappa < \alpha\mathbb{E}[\beta_{k}] .
  \end{cases}
\]
We note that as $\kappa \rightarrow 0$, the update for $\{i=k\}$ becomes equivalent to the $\{i\neq k\}$ case. Optionally, it is possible to update the hyperparameters $\alpha$ and $\kappa$. We place a Gamma prior on $\alpha' = \alpha+\kappa$ and Beta prior on $\kappa' = \frac{\kappa}{\alpha + \kappa}$ as follows:
\begin{align*}
\alpha' \sim& Ga(a^{\alpha},b^{\alpha})\\
\kappa' \sim& Beta(u^{\kappa},v^{\kappa})
\end{align*}
Given the updates for $\alpha'$ and $\kappa'$ it is possible to calculate point estimates for $\alpha$ and $\kappa$.

Before describing the variational updates for $X$ and $S$, we turn our attention to $ELBO$. Using Eq.~\ref{elbo_initial} and Eq.~\ref{joint_log}, we write an equivalent of $ELBO$ up to a constant as
\begin{align}
ELBO \equiv& \mathbb{E} \left [\ln P(S,X,Z \mid \bar{\pi}_{0},\bar{\pi},C,\Theta)\right]\notag\\
&-\mathbb{E} \left [\ln Q(X)\right]  - \mathbb{E} \left [\ln Q(S)\right] \notag\\
&- KL(\bar{\pi}_{0}) -KL(\bar{\pi}) - KL(\bar{\beta})\notag\\
&- KL(C\mid \bar{\beta})- KL(\Theta). \label{compute_ELBO}
\end{align}

\begin{table*}[t]
\caption{Updates for Variational Parameters}
\vskip -0.1in
\label{var_params}
\begin{center}
\begin{sc}
\begin{tabular}{lcccr}
\hline & Initial & Update \\
\hline $u_{i}^{\beta}$ &$1$ &$1 + \sum_{j\neq i}\sum_{j'=1}^{K}\phi_{jj'}(i)$\\[0.5ex]
$v_{i}^{\beta}$ &$\gamma$ &$\gamma + \sum_{i'=i+1}^{K} \sum_{j\neq i'}\sum_{j'=1}^{K}\phi_{jj'}(i')$\\[0.5ex]
$u_{0i}^{\pi}$ &$1$ &$1 + \mathbb{E}[s_{1} = i]$\\[0.5ex]
$v_{0i}^{\pi}$ &$\alpha_{0}$ &$\alpha_{0} + \sum_{i'=i+1}^{K} \mathbb{E}[s_{1}=i']$\\[0.5ex]
$u_{ii'}^{\pi}$ &$1$ &$1 + \sum_{t=2}^{T}\sum_{k=1}^{K}\mathbb{E}[s_{t-1}=i,s_{t}=k]\phi_{ii'}(k) $\\[0.5ex]
$v_{ii'}^{\pi}$ &$\alpha + \kappa$ &$\alpha + \kappa + \sum_{t=2}^{T}\sum_{i''=i'+1}^{K} \sum_{k=1}^{K}\mathbb{E}[s_{t-1}=i,s_{t}=k]\phi_{ii''}(k)$\\[0.5ex]
$a^{\alpha}$ &$\alpha + \kappa$ &$\alpha + \kappa + K^{2}$\\[0.5ex]
$b^{\alpha}$ &$1$ &$1 - \sum_{i=1}^{K}\sum_{j=1}^{K}\mathbb{E}[1-\bar{\pi}_{ij}]$\\[0.5ex]
$u^{\kappa}$ &$\kappa$ &$\kappa+\sum_{i=1}^{K}\sum_{j=1}^{K}\phi_{ij}(i)$\\[0.5ex]
$v^{\kappa}$ &$\alpha$ &$\alpha+\sum_{i=1}^{K}\sum_{j=1}^{K}\sum_{k\neq i}\phi_{ij}(k)$\\[0.5ex]
$\mu_{i}^{\mu}$ &$0$ &$\frac{1}{(b_{\mu})_{i} + N} \mathbb{E}[s_{1}=i]\sum_{n=1}^{N}\mathbb{E}[x_{1}^{n}]$\\[0.5ex]
$\Sigma_{i}^{\mu}$ &$(b_{\mu})_{i}I_{d_{x}}$ & $((b_{\mu})_{i}+N \mathbb{E}[s_{1}=i])I_{d_{x}}$\\[0.5ex]
$a_{i}^{\sigma}$ &$(a_{\Sigma})_{i}$ &$(a_{\Sigma})_{i} + \frac{N}{2}\mathbb{E}[s_{1}=i]$\\[0.5ex]
$b_{i}^{\sigma}$ &$(b_{\Sigma})_{i}$ &$(b_{\Sigma})_{i}+\mathbb{E}[s_{1}=i]\sum_{n=1}^{N}\Sigma_{1}^{n}$\\[0.5ex]
$(\Sigma_{i}^{F})^{-1}$ & $diag(\zeta_{i})$ &$diag(\zeta_{i}) + \sum_{n=1}^{N}\sum_{t=2}^{T}\mathbb{E}[s_{t}=i]\mathbb{E}[(x_{t-1}^{n})^{T}x_{t-1}^{n}]$\\[0.5ex]
$\mu_{i}^{F}$ & $0$ &$(\sum_{n=1}^{N}\sum_{t=2}^{T}\mathbb{E}[s_{t}=i]\mathbb{E}[(x_{t-1}^{n})^{T}x_{t}^{n}])^{T}(\Sigma_{i}^{F})^{-1}$\\[0.5ex]
$(\Sigma_{i}^{H})^{-1}$& $diag(\eta_{i})$&$diag(\eta_{i})+\sum_{n=1}^{N}\sum_{t=1}^{T}\mathbb{E}[s_{t}=i]\mathbb{E}[(x_{t}^{n})^{T}x_{t}^{n}]$ \\[0.5ex]
$\mu_{i}^{H}$ & $0$ &$(\Sigma_{i}^{H})^{-1}\sum_{n=1}^{N}\sum_{t=1}^{T}\mathbb{E}[s_{t}=i]\mathbb{E}[x_{t}^{n}](z_{t}^{n})^{T}$\\[0.5ex]
$a_{id}^{\rho}$ & $a_{i}$ &$a_{i} + \frac{N}{2}\sum_{t=1}^{T}\mathbb{E}[s_{t}=i]$\\[0.5ex]
$b_{id}^{\rho}$ & $b_{i}$ & $b_{i} + \frac{1}{2}\sum_{n=1}^{N}\sum_{t=1}^{T}\mathbb{E}[(z_{t}^{n} - h_{id}x_{t}^{n}\mathbf{1}[s_{t}=i])(z_{t}^{n} - h_{id}x_{t}^{n}\mathbf{1}[s_{t}=i])^{T}]$\\
\hline
\end{tabular}
\end{sc}
\end{center}
\vskip -0.1in
\end{table*}
For the time-invariant parameters we are able to incorporate KL divergence terms. This will prove to be useful in computing $ELBO$ since closed form expressions are readily available for ordinary distributions. Unfortunately, we still need to compute the entropy of $X$ and $S$ separately. In a simple HMM or LDS, there is no need to compute the entropy of time dependent latent variables explicitly. One can simply compute the likelihood of the observation conditioned on the latent variables and the entropy term will cancel out \cite{beal2003variational}. However, in an SLDS time dependent latent variables are tightly coupled and this approach is not applicable; therefore, we need to deal with the first term in $ELBO$. It can be written more explicitly as
\begin{align}
\mathcal{L}'&=\ln P(S,X,Z \mid \bar{\pi}_{0},\bar{\pi},C,\Theta) \notag\\
&= \sum_{i = 1}^{\infty} \mathbf{1}\{s_{1}=i\} \ln\pi_{0i} \notag\\
&+\sum_{t = 2}^{T}\sum_{i = 1}^{\infty} \sum_{i' = 1}^{\infty} \sum_{k = 1}^{\infty} \mathbf{1}\{c_{ii'}=k,s_{t-1}=i,s_{t}=k\} \ln \pi_{ii'}'\notag\\
&-\frac{1}{2}\sum_{n = 1}^{N} \langle\langle(x_{1}^{n}-\mu_{s_{1}}),\Sigma_{s_{1}}^{-1}\rangle\rangle+ \ln | \Sigma_{s_{t}} |\notag\\
&-\frac{1}{2}\sum_{n = 1}^{N}\sum_{t=2}^{T}\langle\langle(x_{t}^{n}-F_{s_{t}}x_{t-1}^{n}), I\rangle\rangle\notag\\
& -\frac{1}{2}\sum_{n = 1}^{N}\sum_{t=1}^{T}\langle\langle(z_{t}^{n}-H_{s_{t}}x_{t}^{n}), R_{s_{t}}^{-1}\rangle\rangle + \ln | R_{s_{t}} | \label{lprime}
\end{align}
where we define $\langle\langle a, B\rangle\rangle \doteq a^{T}Ba$. Using Eq.~\ref{general_var_update} we can write the updates for $Q(X)$ and $Q(S)$ as
\begin{align*}
Q(X) =& \frac{1}{c_{X}}\exp \left\{\mathbb{E}_{Q(S,\Theta)}[\mathcal{L'}]\right\}\notag\\
Q(S) =& \frac{1}{c_{S}}\exp \left\{\mathbb{E}_{Q(\pi_{0},\pi',C,X,\Theta)}[\mathcal{L'}]\right\}
\end{align*}
For these updates it is sufficient to use $\mathcal{L'}$ since other terms in $\mathcal{L}$ only contribute to the constants.

There exist efficient and well studied inference algorithms for LDS models and HMMs such as RTS smoother \cite{rauch1965maximum} and forward-backward algorithm \cite{beal2003variational,murphy2002dynamic}. However, we cannot use the parameters $\Theta$ or $\{\pi,\pi_{0}\}$ directly, this would mean working with $\mathcal{L'}$ instead of its relevant expectation. One solution to this problem is to introduce a set of auxiliary variables to convert $\mathbb{E}_{Q(.)}[\mathcal{L'}]$ into an ordinary likelihood function that can be feed into these algorithms. This approach is first explored in \cite{pavlovic2000impact} and extended in \cite{oh2005variational}.

To update $Q(X)$, first we compute $\lambda_{X} = \left\{\left\{\hat{H}_{t},\hat{R}_{t}\right \}_{t=1}^{T},\left\{ \hat{F}_{t},\hat{U}_{t} \right \}_{t=2}^{T},\hat{\mu},\hat{\Sigma}\right \}$ using Algorithm \ref{alg_lambda_x}, and then we use RTS smoother with $\lambda_{X}$. We note that the expectations in Algorithm \ref{alg_lambda_x} are taken under $Q(S,\Theta)$. With a similar approach we introduce another set of variables $\lambda_{S} = \left\{ \ln\hat{\pi}_{0},\ln\hat{\pi}, \left\{\ln\hat{e}_{t}\right \}_{t=1}^{T}\right \}$ to emulate the likelihood function of an ordinary HMM with $\hat{\pi}_{0}$(the probability of the initial state), $\hat{\pi}$(the probability of transition) and $\hat{e}_{t}$(the probability of emission). We compute $\lambda_{S}$ using Algorithm \ref{alg_lambda_s} and then use forward backward algorithm to infer $Q(S)$. The expectations of the evidence in Algorithm \ref{alg_lambda_s} are taken under $Q(X,\Theta)$. It is usually more convenient to work in log domain to prevent underflow in the forward backward algorithm; therefore, we directly compute the auxiliary variables in log domain. The full variational inference algorithm is given in Algorithm~\ref{vbem}.

\begin{algorithm}[t]
   \caption{Computing $\lambda_{X}$}
   \label{alg_lambda_x}
\begin{algorithmic}
   \FOR{$t=T$ {\bfseries to} $1$}
   \STATE $\hat{R}_{t}^{-1} =
   \mathbb{E}[R_{s_{t}}^{-1}]$
   \STATE $\hat{H}_{t}^{-1} = \hat{R}_{t}
   \mathbb{E}[R_{s_{t}}^{-1}H_{s_{t}}]$
   \STATE $\hat{U}_{t}^{-1} = I +
   \mathbb{E}[\langle\langle H_{s_{t}},
   R_{s_{t}}^{-1}\rangle\rangle ] - \langle\langle
   \hat{H}_{t},\hat{R}_{t}^{-1}\rangle\rangle $
   \IF{$ t\neq T $}
   \STATE $\hat{U}_{t}^{-1} = \hat{U}_{t}^{-1} +
   \mathbb{E}[F_{s_{t+1}}^T F_{s_{t+1}}] -
   \langle\langle \hat{F}_{t+1}, \hat{U}_{t+1}^{-1}\rangle\rangle $
   \ENDIF
   \IF{$ t\neq 1 $}
   \STATE $\hat{F}_{t} = \hat{U}_{t} \mathbb{E}[F_{s_{t}}]$
   \ELSE
   \STATE $\hat{\Sigma}^{-1} = \hat{U}_{t}^{-1}$
   \STATE $\hat{\mu} = \hat{\Sigma}\;
   \mathbb{E}[\Sigma_{s_{1}}^{-1}\mu_{s_{1}}]$
   \ENDIF
   \ENDFOR
\end{algorithmic}
\end{algorithm}
\begin{algorithm}[b]
   \caption{Computing $\lambda_{S}$}
   \label{alg_lambda_s}
\begin{algorithmic}
   \FOR{$i=1$ {\bfseries to} $K$}
   \STATE $\ln\hat{\pi}_{0i} = \mathbb{E}[\ln \pi_{0i}]$
   \FOR{$j=1$ {\bfseries to} $K$}
   \STATE $\hat{\pi}_{ij} = \sum_{k=1}^{K}\mathbb{E}[\pi_{ik}']\phi_{ik}(j)$
   \ENDFOR
   \ENDFOR
   \STATE $\ln\hat{e}_{1}(i) = -\frac{1}{2}\sum_{n=1}^{N}
   \mathbb{E}[\langle\langle x_{1}^{n}-\mu_{i},\Sigma_{i}^{-1}\rangle\rangle
   ] + \mathbb{E}[\langle\langle z_{1}^{n}-H_{i}x_{1}^{n},
   R_{1}^{-1}\rangle\rangle ]+\mathbb{E}[\ln | \Sigma_{i} |] + \mathbb{E}[\ln|
   R_{i} |]$
   \FOR{$t=2$ {\bfseries to} $T$}
   \STATE $\ln\hat{e}_{t}(i) = -\frac{1}{2}\sum_{n=1}^{N}
   \mathbb{E}[\langle\langle x_{t}^{n}-F_{i}x_{t-1}^{n}, I\rangle\rangle ] +
   \mathbb{E}[\langle\langle z_{t}^{n}-H_{i}x_{t}^{n}, R_{i}^{-1}\rangle\rangle
   ]+\mathbb{E}[ln | R_{i} |]$
   \ENDFOR
\end{algorithmic}
\end{algorithm}
\begin{algorithm}[t]
   \caption{VBEM for Sticky HDP-SLDS}
   \label{vbem}
\begin{algorithmic}
   \STATE {\bfseries Initialize:}
   $Q(\bar{\pi}_{0},\bar{\pi},\bar{\beta},C,\Theta,S)$ with hyperparameters
   \REPEAT
   \STATE Compute auxiliary variables $\lambda_{X}$ using Alg.\ref{alg_lambda_x}
   \STATE Infer $Q(X)$ using RTS Smoother
   \STATE Compute entropy of $Q(X)$
   \STATE Compute auxiliary variables $\lambda_{S}$ using Alg.\ref{alg_lambda_s}
   \STATE Infer $Q(S)$ using Forward-Backward
   \STATE Compute entropy of $Q(S)$
   \STATE Update variational parameters as in Table 1
   \STATE Compute KL divergences for
   $\bar{\pi}_{0},\bar{\pi},\bar{\beta},C,\Theta$
   \STATE Compute $\mathbb{E}[\mathcal{L'}]$
   \STATE Compute $ELBO$ as in Eq.~\ref{compute_ELBO}
   \STATE Update hyperparameters
   \UNTIL{$ELBO$ converges}
\end{algorithmic}
\end{algorithm}

\section{Experiments}

We collected data from 10 different subjects using a smarthphone worn around the belt. After a trivial rescaling of the data, we first applied HMM with Gaussian emission probabilities(GaussianHMM). We later used an HMM with GMM as in \cite{trabelsi2013unsupervised}. These two methods are parametric unsupervised method and require the number of modes to be specified ahead of time. Furthermore they do not possess the stickiness property and do not account for autocorrelation. We then compare the HMM model to sticky HDP-SLDS with variational inference. Later we compared the Gibbs Sampler of \cite{fox2008nonparametric} and the proposed variational inference algorithm. Finally we investigated the sensitivity of variational inference to hyperparameter initialization.

\subsection{Data collection}

The experiments are recorded using Galaxy S$4$. Due to the advantage of code written in Android platform, the camera and accelerometer data are recorded synchronously at the rate of approximately $14$ samples per second to be processed offline. The subjects have worn the smartphone around their waist to perform recordings as it is shown in Fig. \ref{fig:GalaxyS4}. The image resolution used for recording is QVGA(320x240). Although the smartphone is capable of recording in higher resolutions, QVGA is selected in order not to increase the computation complexity. The front camera if the device is used to record activities while captured images are displayed on the screen. An illustration of a subject wearing the smartphone on top of the belt can be seen in Fig. \ref{fig:GalaxyS4}.

\begin{figure}[t]
\begin{center}
\end{center}
   \caption{Android Smartphone worn by the subject}
\label{fig:GalaxyS4}
\end{figure}

While wearing the smartphone around the belt, the experimenters performed various activities in indoor environments of a university building and apartment unit. The activities are selected to replicate daily activities in indoor environments for accurate and robust monitoring of a subject. The activities include sitting down, standing up, walking, turning to change direction, going up and down the stairs, and taking the elevator. While performing the activities, accelerometer and images are recorded continuously and simultaneously. The experimenters have performed their daily routine activities during the recordings.

\begin{figure}[t]
\begin{center}
\end{center}
   \caption{Flowchart of data collecting process}
\label{fig:dataRecord}
\end{figure}

While recording the experimental data to be used in training of the proposed model, the procedure is visualized in Fig. \ref{fig:dataRecord}. As it can be observed, 3 dimensional accelerometer data with the gravity component recorded is recorded in sync with the images captured from the camera. For reading the accelerometer data the sample rate of the camera is used since it is slower. Therefore, approximately $14$ samples per second of image and and accelerometer data is captured and recorded. Accelerometer data is recorded as comma separated values(.csv) file whereas the images are recorded into separate folder. Then, $10$ different set of recordings are fed into learning algorithm to derive a model for classification. In the next section, we evaluate the performance of the derived model based on hamming distances and estimated class comparisons with ground truth for activity classes.

\subsection{Metrics}
In parametric models where the number of modes is specified ahead of time, it is common to relabel the predicted sequence and measure the accuracy of the model. The relabeling problem is a combinatorial one and becomes intractable when the number of modes is not small. In non-parametric models though it is more common to use Normalized Mutual Information (NMI). NMI allows us to compare two models resulting in different number of modes. NMI takes values between 0 and 1 where the former corresponds to random assignments. NMI between two collection of sets $\{\mathcal{A}_{i}\}_{i=1}^{N_{\mathcal{A}}}$ and $\{\mathcal{B}_{j}\}_{j=1}^{N_{\mathcal{B}}}$ can be defined as
\begin{align*}
NMI &= \frac{-2\sum_{i}\sum_{j}\frac{\left | \mathcal{A}_{i} \cap \mathcal{B}_{j} \right |}{N}\log\frac{N\left | \mathcal{A}_{i} \cap \mathcal{B}_{j} \right |}{\left | \mathcal{A}_{i} \right |\left | \mathcal{B}_{j} \right |}}{\sum_{i}\frac{\left | \mathcal{A}_{i} \right |}{N}\log\frac{\left | \mathcal{A}_{i} \right |}{N} + \sum_{j}\frac{\left | \mathcal{B}_{j} \right |}{N}\log\frac{\left | \mathcal{B}_{j} \right |}{N}}
\end{align*}

\begin{table}[htb]
\centering
\caption{Comparison of NMI values for Gaussian HMM, Gaussian Mixture HMM and HDP-SLDS}
\label{table:comp_alg}
\begin{tabular}{l|r|r|r|}
\cline{2-4}
                             & \multicolumn{1}{l|}{GaussianHMM} & \multicolumn{1}{l|}{GMMHMM} & \multicolumn{1}{l|}{HDP-SLDS} \\ \hline
\multicolumn{1}{|l|}{Subj01} & 0.156                            & 0.151                       & 0.254                         \\ \hline
\multicolumn{1}{|l|}{Subj02} & 0.214                            & 0.188                       & 0.299                         \\ \hline
\multicolumn{1}{|l|}{Subj03} & 0.161                            & 0.160                       & 0.322                         \\ \hline
\multicolumn{1}{|l|}{Subj04} & 0.188                            & 0.158                       & 0.384                         \\ \hline
\multicolumn{1}{|l|}{Subj05} & 0.168                            & 0.095                       & 0.415                         \\ \hline
\multicolumn{1}{|l|}{Subj06} & 0.246                            & 0.287                       & 0.456                         \\ \hline
\multicolumn{1}{|l|}{Subj07} & 0.156                            & 0.148                       & 0.331                         \\ \hline
\multicolumn{1}{|l|}{Subj08} & 0.206                            & 0.271                       & 0.478                         \\ \hline
\multicolumn{1}{|l|}{Subj09} & 0.168                            & 0.092                       & 0.466                         \\ \hline
\multicolumn{1}{|l|}{Subj10} & 0.206                            & 0.152                       & 0.411                         \\ \hline
\end{tabular}
\end{table}

\begin{figure*}[p]
    \centerline{\includegraphics[width=0.95\textwidth]{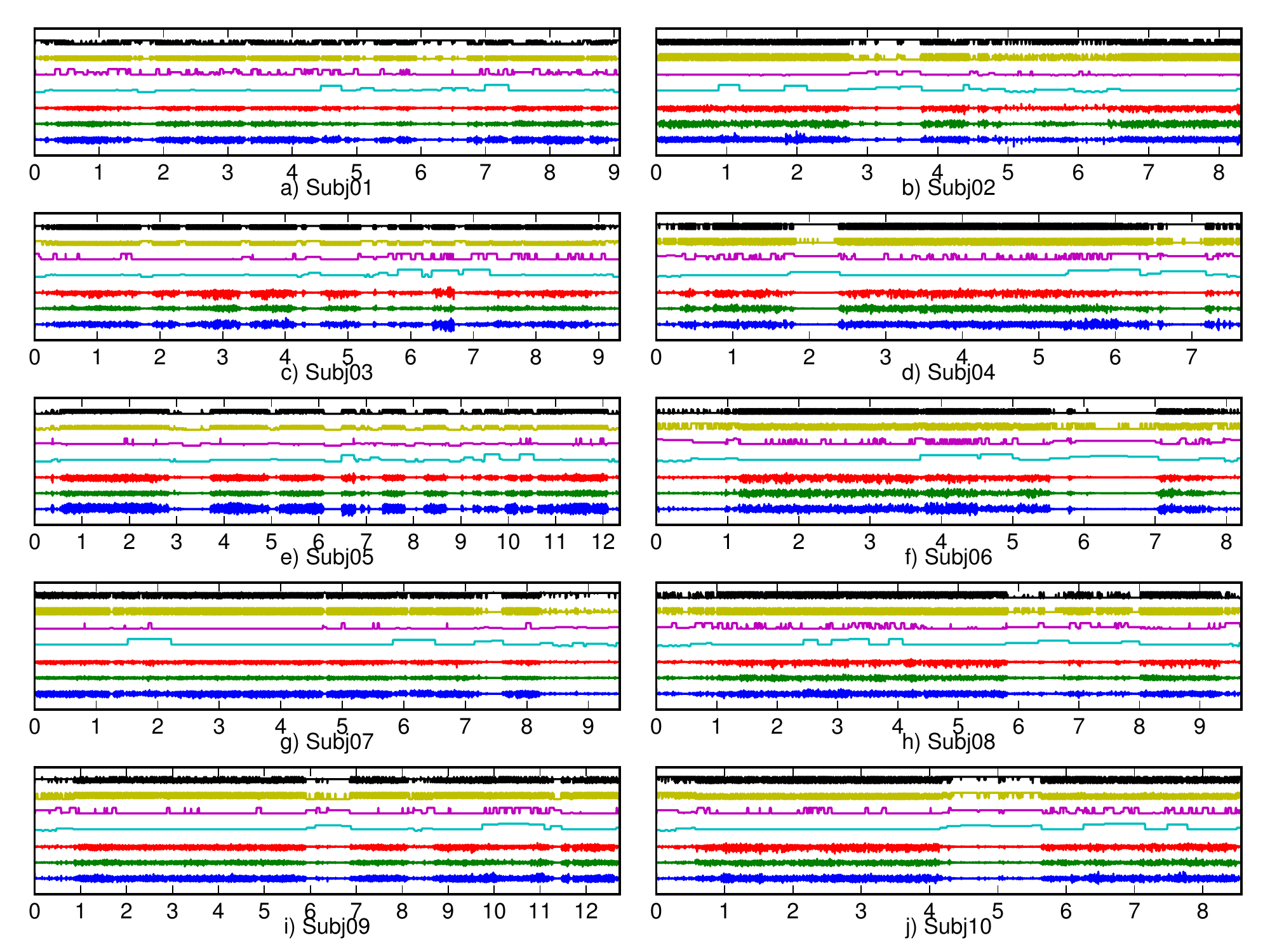}}
    \caption{Raw accelerometer channels along x(blue), y(green) and z(red) directions, true label(cyan) and prediction sequences for sticky HDP-SLDS(pink), Gaussian HMM(yellow) and GMM HMM(black) for all $10$ subjects. The mode indexes for standing, walking, turning, elevator, walking downstairs and walking upstairs are $0,2,3,5,6,7$ respectively.}
    \label{fig:preds}
\end{figure*}

\subsection{Experimental Setup}
We first rescaled each acceloremeter channel such that the standard deviation of the walking mode is $1.0$. This is a trivial rescaling to set the hyper parameters more easily.

In HMM experiments, we provided the true number of modes to the algorithms. The inference is performed via Viterbi and the parameter estimation is done with maximum likelihood. In GMM-HMM, the number of Gaussian Mixtures is set to $3$.

In all sticky HDP-SLDS experiments we fixed the upper limit for the number of modes $K$ to $20$. We first did a grid search in log domain for $\gamma,\alpha_{0},\alpha,\zeta,\eta,\kappa$ after putting uninformative priors to $R,\Sigma$ by setting $a_{\Sigma},a$ to $1.0$ and $b_{\mu},b_{\Sigma},b$ to $100.0$. We then chose $\gamma,\alpha_{0},\alpha,\zeta,\eta,\kappa$ that maximize $ELBO$. Across all subjects, setting $\gamma = 1.0$, $\alpha_{0} = 1.0$, $\alpha = 1.0$, $\zeta = 10.0$, $\eta = 10.0$, $\kappa = 64.0$ provided highest $ELBO$ on average.

We compared Gaussian HMM, GMM-HMM and sticky HDP-SLDS with variational inference. We ran variational inference for $1000$ times and chose the outcome with the highest $ELBO$. Table~\ref{table:comp_alg} summarizes the NMI values for each subject. Figure~\ref{fig:preds} shows the predicted sequences and the true label for three of the subjects.

\begin{figure*}[tp]
    \centerline{\includegraphics[width=0.9\textwidth]{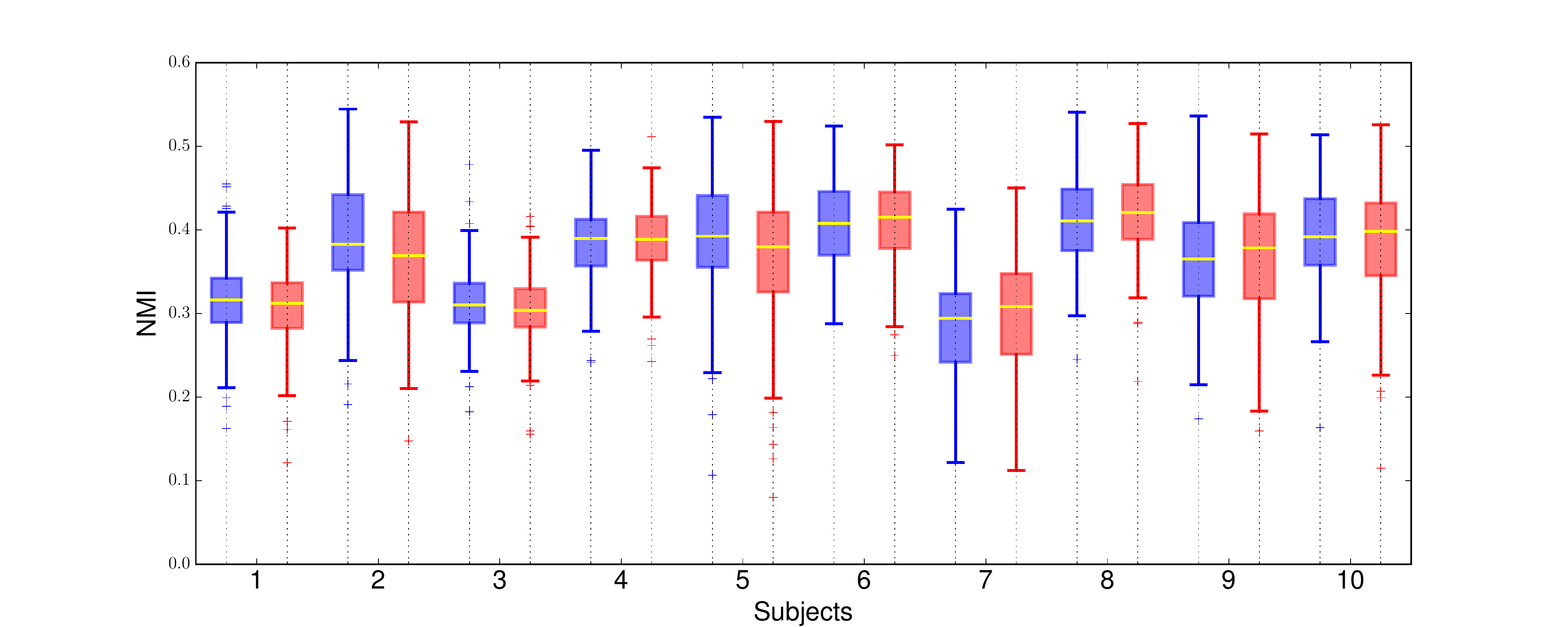}}
    \caption{Box-Whisker plots variational inference(blue) and Gibbs Sampler(red) for 10 subjects}
    \label{fig:comp_vi_and_gibbs}
\end{figure*}

We also compared the variational inference and Gibbs Sampler for sticky HDP-SLDS. Gibbs Sampler is terminated after $100$ iteration. We ran both algorithms for $1000$ times for each subject. Figure \ref{fig:comp_vi_and_gibbs} shows the spread of the NMI values for $100$ trials with the highest $ELBO$ and likelihood for variational inference and Gibbs Sampler respectively. We also measured the cpu time for both algorithms.

Finally, we explored the sensitivity of the model to the initialization of the hyperparameter $b_{\Sigma}$ which is expected to differ most between different modes. Figure \ref{fig:sensitivity} shows the spread for the NMI across $1000$ runs for each subject. In each run $b_{\Sigma}$ is drawn randomly from $[0.125, 0.25, 0.5, 1.0, 2.0, 4.0, 8.0, 16.0, 32.0, 64.0, 126.0]$ for each mode alogn each dimension.

\section{Results}
As seen in Table~\ref{table:comp_alg}, sticky HDP-SLDS outperforms both HMM with Gaussian emission and HMM with Gaussian Mixture emission. There are two reasons for that. First, HMM does not account for the autocorrelation in the signal. Each observation is i.i.d. conditioned on the mode. In SLDS, there is a hidden Markov process $\{x_{t}\}_{t=1}^{T}$ which accounts for the underlying autocorrelation. Secondly, a simple HMM does not have the stickiness property. Therefore it is common to see an oscillatory behavior. Figure~\ref{fig:preds} illustrates this point across all subjects. It is also noteworthy that GaussianHMM outperforms GMMHMM on $8$ of the $10$ subjects. This is in alignment with the assumption that $\{z_{t}\}_{t=1}^{T}$ is drawn from a unimodal distribution. Apart from the quantative and qualitative difference between HMM and sticky HDP-SLDS, we should repeat that HDP-SLDS is a non-parametric model where the number of modes is infered from data.

Overall, we see that walking and standing are captured quite accurately. Even turning and using elevator are captured in most cases. However we failed to distinguish walking upstairs and walking downstairs from walking itself. This is rather eminent in the accelerometer channels. Except for a handful of subjects, the model is able to identify the elevator mode. The model sometimes confuses turning with standing which is a hard problem given the duration of the activity.

Having shown that sticky HDP-SLDS is a better model for the task at hand, we turn our attention to learning problem. Figure~\ref{fig:comp_vi_and_gibbs} depicts the Box-Whisker plots of variational inference and Gibbs sampler for each subject. As expected, the outcomes of both inference methods are consistent. The important difference actually comes in the cpu time. Each round of Gibbs Sampler takes $16.647$ seconds on average whereas variational inference takes $69.784$ seconds on average to converge. To achieve the median nmi value, Gibbs sampler requires $48$ iteration which adds up to $799.056$ seconds on average. In other words, variational inference is $11$ times faster than Gibbs sampler. Given that the most of the time is spent for the inference of temporal parameters $\{x_{t}\}_{t=1}^{T}$ and $\{s_{t}\}_{t=1}^{T}$, the speed becomes especially important. In an online learning setting where the non-temporal parameters are fixed and the temporal parameters are inferred on the fly, it is crucial to finish the inference before the next datapoint arrives. Variational inference can provide this even in low capacity battery sensitive processors of wearable devices.

\begin{figure}[t]
  \centerline{\includegraphics[width=\columnwidth]{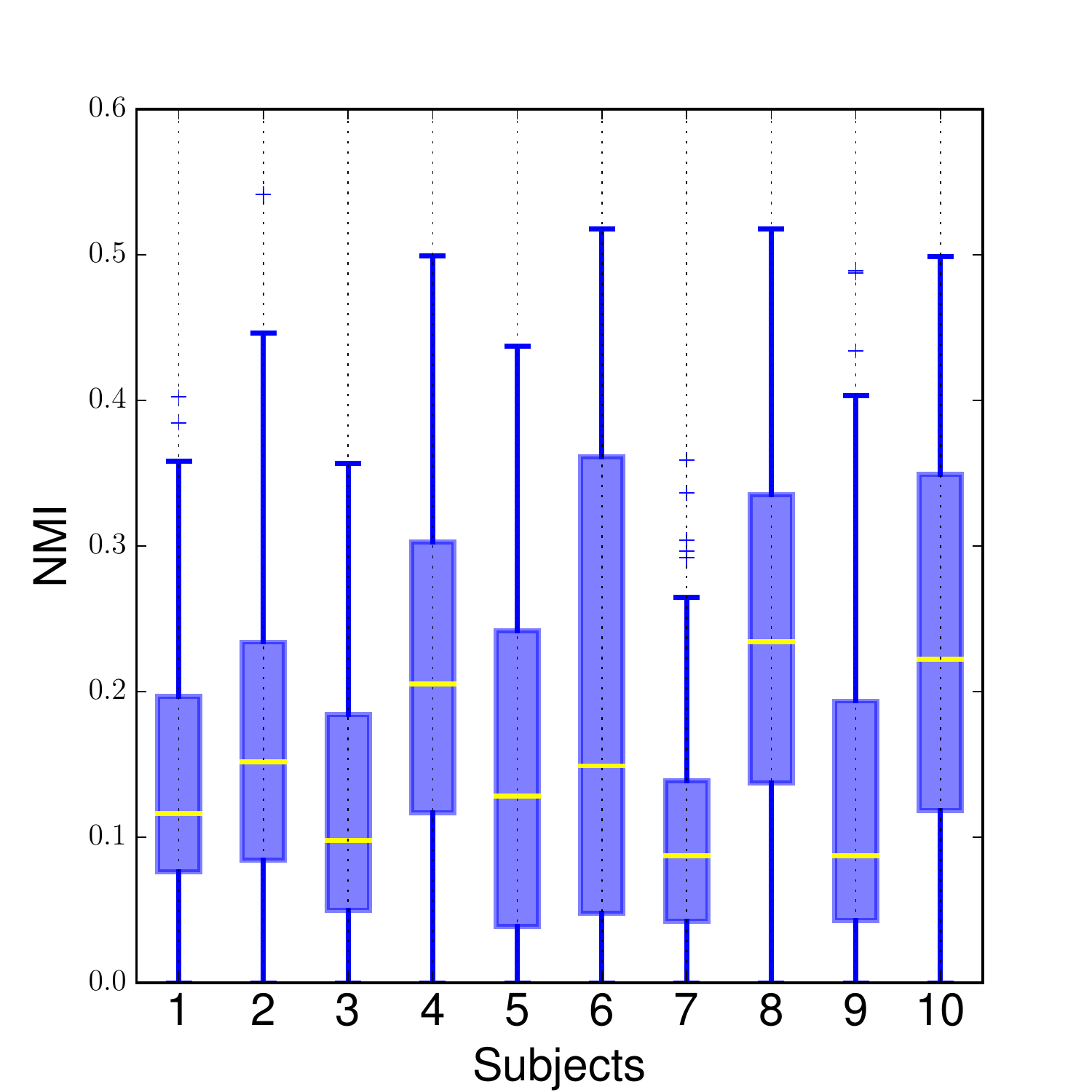}}
  \caption{Box-Whisker plot showing the sensitivity to initialization of $b_{\Sigma}$ for all $10$ subjects}
  \label{fig:sensitivity}
\end{figure}

In activity classification problem, we expect the covariance of the hidden Gaussian process to differ most across modes. We tested the sensitivity of the model fit to the initialization of the hyperparameter $b_{\Sigma}$. A high value for $b_{\Sigma}$ means a lower expected precision therefore a higher variance along that dimension. For an uninformative prior, $b_{\Sigma}$ should be set to a high number. Comparing Figure~\ref{fig:comp_vi_and_gibbs} and Figure~\ref{fig:sensitivity}, it is clear that heavy regularization hurts the performance.

\section{Conclusion}
Inspired by the problem of activity classification with a single acceloremeter, we derived a fast inference method for sticky HDP-SLDS model. Unlike the proposed supervised methods in the literature, HDP-SLDS is capable of learning the activites on the fly. This removes the burden of collecting data for each individual and the restriction on the placement of the smartphone. Furthermore, sticky HDP-SLDS is capable of inferring the number of activities from data. In practice, this translates into the ability to identify activities even when the phone position changes. Compared to the unsupervised approaches in the literature, sticky HDP-SLDS can capture the autocorrelation in the signal and thanks to the stickiness property, it avoids the oscillatory behavior. In theoretival side, our major contribution is the derivation of variational inference for sticky HDP-SLDS. This is the first proposed variational inference for a time series model with stickiness property. We showed that variational inference is an order of magnitude faster than the Gibbs Sampler. This is especially important when the power and the computational resources are limited.


\bibliographystyle{IEEEtran}
\bibliography{ieee_2015_acc}

\begin{IEEEbiography}{Mehmet E. Basbug}
is a PhD candidate at Princeton University Electrical Engineering Department working on machine learning applications with Robert Schapire. He worked on forward stagewise additive models for auto-correlated time series data, Bayesian non-parametric topic models, clustering and factorization without distributional assumptions.
\end{IEEEbiography}

\begin{IEEEbiography}{Koray Ozcan}
(S' 11) received B.S. degree in electrical and electronics engineering with high honors from Bilkent University, Ankara, Turkey, in 2011. He is pursuing Ph.D. degree in the Department of Electrical Engineering and Computer Science at Syracuse University, Syracuse, NY. His research interests include image processing, embedded systems, and computer vision.
\end{IEEEbiography}

\begin{IEEEbiography}{Burak Kakillioglu}
received B.S. degree in electrical and electronics engineering from Bilkent University, Ankara, Turkey in 2015. He is Ph.D. candidate at Department of Electrical Engineering and Computer Science in Syracuse University and he is working as a research assistant in Smart Vision Systems Laboratory at the same department. His research interests are computer vision, artificial intelligence, and network systems.
\end{IEEEbiography}

\begin{IEEEbiography}{Senem Velipasalar} (M'04) received the Ph.D. and M.A. degrees in electrical engineering from Princeton University, Princeton, NJ, in 2007 and 2004, respectively, the M.S. degree in electrical sciences and computer engineering from Brown University, Providence, RI, in 2001, and the B.S. degree in electrical and electronic engineering with high honors from Bogazici University, Istanbul, Turkey in 1999.

During the summers of 2001 to 2005, she was with the Exploratory Computer Vision Group, IBM T. J. Watson Research Center, Yorktown Heights, NY. Between 2007 and 2011, she was an Assistant Professor with the Department of Electrical Engineering, University of Nebraska-Lincoln, Lincoln. Currently, she is an Assistant Professor with the Department of Electrical Engineering and Computer Science, Syracuse University, Syracuse, NY.  The focus of her research has been on wireless embedded smart cameras, multicamera tracking and surveillance systems, and automatic event detection from videos. Her current research interests include embedded computer vision, video/image processing, embedded smart camera systems, distributed multi-camera systems, pattern recognition and signal processing.

Dr. Velipasalar received a Faculty Early Career Development Award (CAREER) from the National Science Foundation (NSF) in 2011. She is the coauthor of the paper, which received the 3rd place award at the 2011 ACM/IEEE International Conference on Distributed Smart Cameras. She received the Best Student Paper Award at the IEEE International Conference on Multimedia and Expo in 2006. She is the recipient of the EPSCoR First Award, two Layman Awards, the IBM Patent Application Award, and Princeton and Brown University Graduate Fellowships.
Biography text here.
\end{IEEEbiography}


\end{document}